\documentclass[lettersize,journal]{IEEEtran}
\usepackage{amsmath,amsfonts}
\usepackage{algorithmic}
\usepackage[T1]{fontenc}  
\usepackage[utf8]{inputenc}  

\usepackage{booktabs}
\usepackage{algorithm}
\usepackage{algorithmic}
\usepackage{array}
\usepackage{xcolor}
\usepackage{amssymb}
\usepackage[caption=false,font=normalsize,labelfont=sf,textfont=sf]{subfig}
\usepackage{textcomp}
\usepackage{stfloats}
\usepackage{url}
\usepackage{verbatim}
\usepackage{graphicx}
\usepackage{cite}
\begin{document}

\title{Neuroplasticity in Artificial Intelligence -- An Overview and Inspirations on Drop In \& Out Learning}
\author{Yupei Li, Manuel Milling, Björn W. Schuller,~\IEEEmembership{Fellow,~IEEE}
\thanks{Yupei Li, GLAM, Imperial College London, London, UK, email: yupei.li22@imperial.ac.uk}
\thanks{Manuel Milling, CHI, Technical University of Munich, Munich, Germany, email: manuel.milling@tum.de}
\thanks{Björn W. Schuller, GLAM / CHI, Imperial College London / Technical University of Munich, London, UK / Munich, Germany, email: schuller@tum.de}
\thanks{This research was partially supported by the Munich Center for Machine Learning and the Munich Data Science Institute.}
}

\markboth{Journal of \LaTeX\ Class Files}%
{Shell \MakeLowercase{\textit{et al.}}: Neurogenesis and Neuroplasticity in Deep Neural Networks:  An Overview and Inspirations on Drop In \& Out Learning}

\IEEEpubid{
  \parbox{\textwidth}{ 
    \centering 
    ~\copyright~2025 The Authors. This work is licensed under a Creative Commons Attribution-NonCommercial-NoDerivatives 4.0 License.\\
    For more information, see \url{https://creativecommons.org/licenses/by-nc-nd/4.0/}
  }
}



\maketitle
\begin{abstract}
Artificial Intelligence (AI) has achieved new levels of performance and spread in public usage with the rise of 
deep neural networks (DNNs).
Initially inspired by human neurons and their connections, NNs have become the foundation of AI models for many advanced architectures. 
However, some of the most integral processes in the human brain, particularly neurogenesis and neuroplasticity in addition to the more spread neuroapoptosis have largely been ignored in DNN architecture design.
Instead, contemporary AI development predominantly focuses on constructing advanced frameworks, such as large language models, which retain a static structure of neural connections during training and inference.
In this light, we explore how neurogenesis, neuroapoptosis, and neuroplasticity can inspire future AI advances. 
Specifically, we examine analogous activities in artificial NNs, introducing the concepts of ``dropin'' for neurogenesis and revisiting ``dropout'' and structural pruning for neuroapoptosis. We additionally suggest neuroplasticity combining the two for future large NNs in ``life-long learning'' settings following the biological inspiration. We conclude by advocating for greater research efforts in this interdisciplinary domain and identifying promising directions for future exploration.
\end{abstract}


\section{Introduction}
\IEEEPARstart{N}{eural} networks (NNs) have been extensively developed and demonstrated to be effective in recent years, building upon their initial inspiration from the functioning of human brain neurons \cite{mcculloch1943logical}. 
The rapid recent advancements in artificial intelligence (AI) have been driven by the foundational principles and efficacy of NNs, which have garnered important attention for the development of cutting-edge architectures, such as deep NNs (DNNs), Transformers \cite{vaswani2023attentionneed}, and large models such as large language models (LLMs) built upon these. Moreoever, there are numerous techniques available to ensure the high performance of such models, including those that enhance generalisation and robustness. 
Since these techniques are all built upon NNs, which are inspired by the human brain, we contend that certain aspects of novel techniques are also influenced by the activity of brain neurons.

In biology, the human brain follows a lifelong cycle, during which it undergoes three important stage: neurogenesis, neuroapoptosis, and dynamic neuroplasticity throughout its life. 
\textbf{Neurogenesis}, the process of generating new neurons, persists in specific regions of the adult mammalian brain \cite{altman1962new, gotz2005cell, gage2002neurogenesis}. Neurogenesis 
is the process of the production of new neurons. When the neurons' life cycle terminates, neurons undergo \textbf{neuroapoptosis} \cite{bredesen1995neural, yuan2000apoptosis}, which represents an intrinsic self-destruction program that leads to the elimination of the neurons. During their whole life, once generated, these neurons are subject to stimulation from both internal mechanisms and the external environment, which leads to another important process: \textbf{neuroplasticity}. This is commonly defined as the brain's ability to reorganise itself by forming new neuronal connections or through cortical remapping in response to experience \cite{perwej2012neuroplasticity}. More broadly, neuroplasticity encompasses both neurogenesis and neuroapoptosis, as they are considered subtypes of neuroplasticity when their outcomes contribute to the brain's capacity for reorganization. However, neuroplasticity involves additional mechanisms, such as synaptic plasticity, which facilitates the adjustment of neural connections; functional plasticity, including compensatory plasticity, whereby one brain region takes over the function of another that has been impaired; and structural changes in the brain, such as the growth of dendrites to support the formation of new synapses.
Various alternative definitions of neuroplasticity have been explored in previous research but they share similar ideas \cite{demarin2014neuroplasticity, quartarone2022defining, wenger2021neuroplasticity}.

\IEEEpubidadjcol

Humans possess the physiological capacity to form new neural connections while eliminating redundant neurons in response to environmental changes and brain injuries. This process facilitates functional recovery by rerouting tasks through intact neural pathways and enables learning \cite{Berger_2019, galvan2010neural}. Inspired by this adaptive capability of the brain, similar strategies have been observed in \textit{artificial} NNs, which encounter analogous stimuli in real-world applications. For instance, the continuous influx of data often necessitates the restructuring of neural networks, akin to the human brain encountering new learning tasks that require neuroplasticity.
The scenarios of external stimuli in biological learning processes also have an analogy in NNs. For instance, in reinforcement learning, rewards from the environment play a similar role.

A well-known example of existing strategies that resemble brain mechanisms
is \emph{``dropout''} \cite{srivastava2014dropout}, 
similar to \emph{neuron apoptosis} in 
a mammalian
brain, a regularisation technique that largely improves the performance of NN-based architectures reducing overfitting by randomly deleting neurons, as not all neurons are required for specific tasks. This phenomenon parallels part of \emph{neuroplasticity} in the human brain. Conversely, we introduce the concept of \emph{``dropin''}, the opposite of dropout, which involves adding new neurons to existing NNs. This process is analogous to \emph{neurogenesis} and holds important potential, as increasing the number of parameters can enhance model capacity, as demonstrated by LLMs achieving state-of-the-art performance in oceans of applications.

In this light, this paper aims to address: 
\begin{itemize}
    \item Providing an overview of the motivation and existing literature on
    neurogenesis, 
    neuroapoptosis, and neuroplasticity in both biology and Deep Learning.
    \item Provide an overview of the existing literature on dropout, propose a definition of ``dropin'', and review similar techniques.
    \item Propose algorithms for 
    artificial 
    neuroplasticity.
    \item Offering a discussion on the implications, future challenges, and potential research directions.
\end{itemize}

\section{Neuroplasticity in the Human Brain}
As previously mentioned, the dynamic adjustment of neurons and their connections is crucial in the human brain. The brain frequently employs mechanisms such as neurogenesis, neuroapoptosis, and  neuroplasticity to achieve this adaptability. These processes occur on a daily basis in humans, with their frequency influenced by individual health conditions. The human brain contains approximately 86 billion neurons \cite{10.1093/brain/awae390}, and around 700 new neurons are generated through neurogenesis in adults' hippocampus \cite{spalding2013dynamics} 
every day,
although precise estimations for neuroplasticity \cite{cunnington2019neuroplasticity} and neuroapoptosis \cite{yuan2000apoptosis} remain elusive. 

These processes can occur for various reasons. Neurogenesis is influenced by factors such as embryonic development and brain maturation \cite{urban2014neurogenesis}, external stimuli that stimulate hippocampal generation \cite{kempermann2015neurogenesis}, or pathological conditions during disease development \cite{kaneko2009adult}, and further more. Neuroapoptosis also occurs through two main mechanisms: ligation of the tumor necrosis factor (TNF) family receptor, which activates the caspase pathway, and trophic factor deprivation, DNA damage, or endoplasmic reticulum stress, which trigger neurodegenerative changes \cite{PALIWAL2022165}. Neuroplasticity can be driven by structural changes in the brain, functional plasticity in response to external stimuli \cite{kolassa2007structural}, or compensatory plasticity in the event of damage \cite{frykberg2015neuroplasticity}, and further more. Although these processes occur at the micro level within the brain, human behaviour can influence them. For instance, factors such as diet and exercise can affect brain function. Specifically, a diet rich in curcumin, known for its antioxidant properties, may help prevent the rapid aging of the brain, while exercise that promotes the production of brain-derived neurotrophic factor can contribute to the expansion of brain size \cite{maharjan2020role}.

These processes are essential for the normal functioning of daily life, providing important benefits to the brain. The generation of neurons and connections is crucial for memory formation and learning \cite{deng2010new}, with evidence suggesting that newborn neurons contribute differently at various stages of maturation. Additionally, neurogenesis plays a role in mood regulation \cite{anacker2017adult}, as research has highlighted the importance of dentate gyrus function in influencing emotions. Furthermore, it enhances cognitive ability \cite{martinez2024hippocampal} by facilitating the degradation of established memories and promoting relearning.

The selective reduction of certain neurons and connections also serves important functions. It contributes to neural efficiency and may even aid in the treatment of certain diseases \cite{fornito2015connectomics}, which suggests that degeneracy of neurons enables adaptive responses. Moreover, neuronal refinement eliminates excess neurons, ensuring the selection of essential ones \cite{lindenberger2019brain}. These mechanisms are fundamental and indispensable for optimal brain function.

\section{The Quest for Optimal NN configurations}
DNNs are inspired by biological neurons; however, in AI, the primary focus is on designing optimal networks to achieve the best possible performance. 
The fundamental challenge in DNN design is to find a neural network configuration that best solves a given task. 
The Universal Approximation Theorem \cite{hornik1989multilayer} states that, in theory, a neural network can approximate any function given sufficient capacity. However, in practice, in addition to the challenge of optimising the weights of a given NN through training, the initial problem comes in the form of the architectural NN design. It involves a trade-off between training cost and model capacity under consideration of task as well as type and amount of training data, which can lead to issues such as overfitting or underfitting. 
In other words, the optimal number of parameters in an NN cannot be determined analytically, mirroring the inherent uncertainty in defining the precise number of neurons required in the human brain. While some studies propose default starting points \cite{hunter2012selection} or approximate parameter ranges \cite{nguyen2023neuronsneedrefinedanalysis, berngardt2024minimumnumberneuronsfully}, the network architecture must still be dynamically adjusted based on the task environment.

\subsection{Duality of Small and Large Networks} 
Without a doubt, the success story of NNs -- and AI in general -- has largely been written by increasing the depth, or more generally the size of NNs with AlexNet \cite{krizhevsky2012imagenet} as an example. This trend has accelerated in recent years fueled by LLMs and revealed new insights into NN learning behaviour as, for instance, double descent \cite{schaeffer2023doubledescentdemystifiedidentifying}.
Nevertheless, at least for the sake of computational efficiency, designs of large and powerful networks have always inspired the search for smaller network architectures aiming to achieve on-par performance, where Squeezenet \cite{iandola2016squeezenetalexnetlevelaccuracy50x} and efficientnet \cite{tan2020efficientnetrethinkingmodelscaling} are notable examples.

\paragraph{Lottery ticket hypothesis}
On a more fundamental level, this quest for smaller networks is supported by the lottery ticket hypothesis (LTH), proposed by Frankle et al.\ \cite{frankle2019lotterytickethypothesisfinding}. It states that
within randomly initialised large-scale NNs, there exists a small-scale sub-network with effective weights, capable of being trained to achieve performance comparable to the entire large-scale NN. These sub-networks are referred to as lottery NNs. Moreover, it strongly suggests that not all components within an NN are essential. Zilly et al.\ \cite{zilly2022plasticity} also share similar ideas on this.

This hypothesis has also been validated in previous literature. Liu et al.\ \cite{liu2024surveylotterytickethypothesis} conducted a survey on models incorporating this hypothesis and discussed existing challenges, such as acceleration and its application to diffusion models. Specifically, experiments have been designed to verify this hypothesis across various deep neural networks, including convolutional neural networks (CNNs) \cite{da2022proving}, spiking neural networks (SNNs) \cite{kim2022exploring}, and graph-based neural networks (GNNs) \cite{wang2023graph}. Additionally, Transformer-based architectures, such as the Bidirectional Encoder Representations from Transformers (BERT) \cite{chen2020lottery} and Vision Transformers (ViTs) \cite{shen2022data}, have also been examined in this context. 
Nevertheless, even under the assumption of the existence of a winning ticket, the LTH does not provide a straight forward approach of how to retain such a best small-scale NN with the highest performance.

\paragraph{Student-teacher model}
Student-teacher models on the other hand belong to the more applicable attempts to reduce model size while keeping high performance.
The idea was proposed by Hinton et al.\ \cite{hinton2015distilling}, where a well-trained large model serves as the ``teacher model'' and a less complex model acts as the ``student model''. There are two main directions in the student-teacher framework. The first is knowledge distillation, in which the teacher transfers its knowledge to the student; Hu et al.\ \cite{hu2023teacher} conducted a survey on this approach. The second is reverse knowledge distillation, where the student model transfers its acquired knowledge back to the teacher to enhance its capabilities \cite{nasser2024reverse}. These two approaches constitute the ``learning-to-teach'' paradigm \cite{fan2018learningteach}. 

The knowledge distillation process can be interpreted in relation to the concepts of neuroplasticity, as it can be likened to the identification and dynamic removal of useless knowledge from neurons in large-scale teacher models. This results in smaller-scale student models with concentrated and selectively retained knowledge. It demonstrates the advantages of neural networks of different sizes, through which distilled knowledge flows. In contrast, the reverse process differs, involving the student model providing additional information to the teacher for verification.

\subsection{Dynamic tasks solution}
The concepts introduced up until this point emphasise that the non-trivial challenges within the search for well-suited neural network configurations for any task. 
However, these challenges are further magnified if the task is dynamically adjusted or its environment changes over time.
These conditions are a central focus in continuous learning, where models must adapt dynamically to evolving conditions while retaining previously learnt knowledge.
In this case, the learnt states of NNs need to be updated over time under the consideration of new data points, thereby overcoming problems such as catastrophic forgetting.

Continuous learning has relations to student-teacher model as aforementioned. Lee et al.\ \cite{lee2021continual} proposed the use of multiple specialised teachers to introduce distinct features to students, thereby facilitating continuous learning. Similarly, Ye et al.\  \cite{ye2021lifelong} framed this as a lifelong learning framework, based on experiments with Generative Adversarial Networks (GANs) and Variational Autoencoders (VAEs). Additionally, Lu et al.\ demonstrated specific applications in target identification \cite{lu2021continuous}. These studies highlight the effectiveness of student-teacher models and underscore the potential of continuous learning and neuroplasticity.

Moreoever, continuous learning can also be viewed as lifelong learning \cite{chen2018lifelong}, wherein trained NNs can be fine-tuned and adapted in response to environmental changes, such as dataset influxes and application shifts. Continuous learning can relate to the dynamic modification of NN structures to optimise performance in evolving environments and changing requirements \cite{liu2017lifelong, billiot2023continuous}. NNs must be updated over time through incremental learning while addressing the challenge of catastrophic forgetting \cite{Parisi2019}, as discussed in more detail below.

Geng et al. \cite{Geng2009} defined incremental learning as the process of learning with the influx of new samples, while Van et al. \cite{van2022three} surveyed three types of incremental learning: task-incremental \cite{pourkeshavarzi2021looking}, domain-incremental \cite{mirza2022efficient, lamers2023clustering}, and class-incremental learning \cite{mittal2021essentials}. These studies highlight the effectiveness of incremental learning across various algorithms and applications. Notably, LLMs also employ techniques such as Low-Rank Adaptation (LoRA) \cite{hu2021loralowrankadaptationlarge} to adapt to new tasks, demonstrating their ability to learn and update their capabilities based on new data.

The term catastrophic forgetting describes that 
models tend to forget information that is less relevant to current tasks in sequential learning, and was first identified by McCloskey et al.\  \cite{mccloskey1989catastrophic}. 
Chen et al.\ \cite{chen2018continual} linked this issue to continuous learning, noting that it is a challenge to address. In contrast, Shintre et al.\ \cite{shintre2019making} argued that intentionally forgetting certain content can be beneficial for data protection. Sha et al.\ \cite{sha2024forgettingmachinelearningbeyond} proposed that forgetting could also help mitigate overfitting and address memory constraints. Striking a balance between what to forget and what to retain represents a key research challenge, with ongoing work focused on defying forgetting \cite{de2021continual} and enhancing the forgetting process \cite{ullah2009you}. 

There are multiple methods and algorithms and could be splitted as follows.
\paragraph{Online Learning}
Hoi et al.\ \cite{hoi2018onlinelearningcomprehensivesurvey} provide a review of online learning strategies by discussing methods for learning from a sequence of data instances one at a time to enhance model capabilities. Liu et al.\ \cite{liu2020learning} explored how online learning provides AI with new opportunities to interact with humans and the environment, enabling continuous learning of new tasks. 

Several AI models have emerged that utilise online learning methods. Perez et al.\ \cite{perez2018review} surveyed online learning techniques primarily based on adaptive learning for NNs, while Jain et al.\ \cite{jain2014review} focused on its applications in supervised learning. Early investigations, such as those by Eon et al.\ \cite{eon1998online}, proposed an online learning framework using stochastic approximation theory. Building on this, some DNNs have been adapted for online learning. Ergen et al.\ \cite{ergen2017efficient} discussed effective online learning methods using particle filtering (PF) to update long short-term memory (LSTM) parameters. Marschall et al.\ \cite{marschall2020unified} expanded on this topic by explaining a framework for recurrent NNs. Sahoo et al.\ \cite{sahoo2017onlinedeeplearninglearning} introduced a novel framework employing the Hedge Backpropagation (HBP) method for online deep learning. More recently, LLMs have also benefited from online learning \cite{hao2024online}.

In terms of specific applications, Cui et al.\ \cite{cui2016continuous} explored online learning within the context of unsupervised learning, while Choy et al.\ \cite{choy2006neural} proposed a multistage online learning process to solve control problems. Recent applications include agent conversation systems \cite{dai2025multi} and agent-based web learning \cite{qi2024webrl}, and further more. There is a growing body of work focused on online learning, demonstrating its effectiveness in dynamically adjusting the knowledge base of NNs.
\paragraph{Transfer Learning}
Several surveys have been conducted on this topic \cite{weiss2016survey, torrey2010transfer, zhuang2020comprehensive}, concluding that transfer learning is defined as the process of training models with data from a different but related domain for pre-training, and subsequently transferring the acquired knowledge to the target application. Benavides-Prado et al.\ \cite{benavidesprado2022theoryknowledgetransfercontinual} propose a formal theory of knowledge transfer in transfer learning and provide a detailed mathematical proof. Meanwhile, Khodaee et al.\  \cite{khodaee2024knowledge} systematically review knowledge transfer learning in the context of lifelong continuous learning. 

Several surveys have examined transfer learning in various domains, including computer vision (CV) \cite{shin2016deep}, natural language processing (NLP) \cite{alyafeai2020survey}, speech processing \cite{wang2015transfer}, and DNN applications more broadly \cite{tan2018survey}, among others. Various approaches have been proposed to implement transfer learning, with fine-tuning being one of the most widely used techniques. Differential Evolution-based adaptive fine-tuning has been applied to CNNs, achieving superior performance on classification tasks \cite{vrbanvcivc2020transfer}. Additionally, SpotTune has been introduced to automatically determine whether a given sample should be processed through pre-training or fine-tuning layers, allowing for adaptive adjustments \cite{guo2019spottune}.

Another prominent approach is domain adaptation, which has been extensively reviewed in prior surveys \cite{kouw2018introduction, kamath2019transfer}. For instance, CNNs have been adapted for skin disease classification through a two-step progressive transfer learning technique applied to a specialised dataset \cite{gu2019progressive}. Visual domain adaptation (VDA) has also been proposed to transfer knowledge for image classification tasks.

Beyond these methods, zero-shot and few-shot learning are widely recognised alongside transfer learning. Current LLMs have successfully incorporated these strategies, achieving remarkable generalisation capabilities \cite{tong2024t3, yue2024less}.
Empirical studies have demonstrated that techniques such as pruning CNNs \cite{molchanov2016pruning} and parameter-efficient fine-tuning with Adapter modules \cite{houlsby2019parameter} can largely reduce the number of parameters while maintaining or even improving efficiency.
\paragraph{Reinforcement Learning}
The literature suggests that reinforcement learning is another approach to continuous learning, as its definition inherently involves an infinite sequence of reward collection, with policies dynamically adapting based on feedback \cite{abel2023definitioncontinualreinforcementlearning}. Khetarpal et al.\ \cite{khetarpal2022towards} provide a comprehensive summary of continuous reinforcement learning and propose future directions related to neuroscience. Notably, they discuss human reward mechanisms inspired by the human brain, which could aid in designing reward functions by offering insights into the origins of rewards. Additionally, they highlight the stability-plasticity trade-off in continuous learning. Several reinforcement learning methods also emphasise their connection to continuous learning strategies. For instance, reinforcement learning has been used to expand neural network layers dynamically for new tasks \cite{xu2018reinforced}.


In short, most approaches that operate in the continuous learning problem are focused around a dynamical adaptation of NN weights and leave the NN architecture constant over time.
In our opinion, this leaves some room for improvement as an ``optimal'' or at least well-suited NN architecture and thus model complexity might change as well over time, as for instance more fine-grained classes are added to a task.
As a consquence, we hypothesise that inspirations from the neuroplasticity of the human brain may manifest themselves to dynamically changing network architectures, which ultimately overcome some challenges of continuous learning.

\section{Neurogenesis, Neuroapoptosis, and Neuroplasticity in AI}
\label{sec1}

Recent literature establishes connections and elucidates the interface between aforementioned biological concepts and AI. Sadegh et al.\ \cite{sadegh2024neural} explore the analogous principles between neural structures in the brain and their advancement for AI development, emphasising NN weights as representations of synaptic transmission efficiencies. Saxena et al.\ \cite{saxena2024bridging} further intertwine these domains through the concept of environmental enrichment, positing that environmental rewards stimulate brain neurons and similarly enhance NNs' capabilities through transfer learning. 
These studies lay foundational groundwork for ongoing research in this interdisciplinary field, but lack concrete applications of neuroplasticity to artificial NNs.
To paint a picture on what might fill in this gap, we break down in the following neuroplasticity into a combination of neurogenesis, neuroapoptosis. We outline already existing and promising avenues towards artificial pendents to these individual components as well as the combined neuroplasticity.

\subsection{Neurogenesis: proliferation and dropin}
Some ANN research has further taken inspiration from neurogenesis to grow the amount of neurons in an architecture over time. 
Huang et al.\ \cite{huang2023neurogenesis} propose a novel drop-and-grow strategy that progressively reduces the number of non-zero weights while maintaining extremely high sparsity and high accuracy. This approach involves dropping weights close to zero and replacing them with newly generated ones, aligning more effectively with the core principles of Spiking Neural Networks (SNNs) for improved biological plausibility and efficiency. Sowrirajan et al.\ \cite{sowrirajan2024enhancing} propose that neurogenesis is triggered only when the subsequent accuracy is less than or equal to the previous accuracy. Additionally, the emergence of new neurons is predicted using the Mamdani fuzzy inference system. In reinforcement learning, Eriksson et al.\ \cite{eriksson2019dynamic} propose a method where errors exceeding a predefined surprise threshold are interpreted as encounters with unknown experiences. In response, new nodes are directly added to the network to adapt to novel situations. 
Given the scarcity of research on neurogenesis, we suggest exploring a new avenue of ``drop-in" learning, as defined below. Inspired by biological brain neurogenesis, we define neurogenesis in artificial NNs as \textbf{the formation of new neurons or connections between neurons following a technical approximation.}

\subsubsection{Dropin}
\label{dropin}
We consider dropin to be the conceptual opposite of dropout, analogous to neurogenesis in biological brain systems. 
We define dropin here as \textbf{an augmentative technique that randomly activates additional neurons and connections during training, increasing the network's ability to learn and enhancing its capacity to develop robust representations by dynamically adapting to increased complexity}.

Inspired by pruning algorithms and mechanisms within the brain, we define the ``dropin" strategy, which involves activating and selecting layers for additional capacity based on two conditions. First, it could be triggered once by an environment change, such as new tasks or data influx. Second, it could be activated if the NN has reached its maximum capacity but still fails to meet practical requirements. Specifically, activation occurs when a predefined criterion is met; we take convergence of the model as a sign that a NN reaches its capacity formulated in Equation \ref{eq:reach},
\begin{equation}
\label{eq:reach}
    \Delta loss = |loss_{epoch} - loss_{epoch-1}| < \delta
\end{equation}
where $\delta$ is pre-defined threshold. The corresponding algorithm is presented in Algorithm \ref{al:dropin}.

\begin{algorithm}
\caption{Dropin Process in NNs}
\label{al:dropin}
\begin{algorithmic}[1]
\REQUIRE Input $\mathbf{x}$, weight matrices $\mathbf{W}^{(l)}$ for each layer $l$, $Criterion$ for measuring the performance
\ENSURE Output $\mathbf{y}$

\STATE \textbf{Training Phase:}
\FOR{New data \OR $\Delta \max(\text{loss}_{epoch}, \text{loss}_{epoch-1}) < \delta$}

\item  \COMMENT{The layer with the largest LRP value exceeds the threshold defined by the average LRP value.}
        \STATE Sample a number of new neurons and connections for layer $i$ $\mathbf{W}^{(l)_{new}}$ \COMMENT{Create additional neurons}
        \STATE Compute activations:

        $\mathbf{h}^{(l-1)_{new}} = f(\text{concat}(\mathbf{W}^{(l-1)},\mathbf{W}^{(l-1)_{new}})  \mathbf{h}^{(l-2)})$

        $\mathbf{h}^{(l)} = f(\text{concat}(\mathbf{W}^{(l)},\mathbf{W}^{(l)_{new}})  \mathbf{h}^{(l-1)_{new}})$ \COMMENT{Two consecutive layers will be influenced as the shape of weight matrices has been changed}

    \IF{$Criterion(val)$ consistently decrease}
    \STATE \textbf{break} 
    \COMMENT{Check validation after training stage}
\ENDIF
\ENDFOR
\STATE Output $\mathbf{y} = \mathbf{h}^{(L)}$

\STATE \textbf{Inference Phase:}
\FOR{each layer $l$}
    \STATE Compute activations: $\mathbf{h}^{(l)} = f(\mathbf{W}^{(l)} \mathbf{h}^{(l-1)})$ \COMMENT{Use all neurons during inference}
\ENDFOR
\STATE Output $\mathbf{y} = \mathbf{h}^{(L)}$

\end{algorithmic}
\end{algorithm}

Despite no clear mentioning of the described approach in literature -- to the best of the authors' knowledge -- our suggested technique of dropin learning shares fundamental similarities to approaches such as model expansion and adaptive learning, which we further discuss in the following.

\paragraph{Model Expansion}
Recent advances in model expansion focus on enlarging neural architectures to incorporate additional task-specific features or mitigate underfitting. Wu et al.\ \cite{wu2024llama} extend LLaMA through block expansion using only new corpora, a widely recognised training strategy in LLMs known as post-training. Additionally, research on smaller models remains active; for instance, Amiriparian et al.\ \cite{amiriparian2024exhubert} expand entire layers for emotion detection tasks, achieving notable performance improvements.

\paragraph{Adaptive Learning}
Adaptive learning has gained increasing attention, particularly in domain-specific fine-tuning and post-training of LLMs. Mundra et al.\ \cite{mundra2024comprehensive} provide a comprehensive overview of current adapter-based approaches and their challenges, highlighting their role as a parameter-efficient fine-tuning (PEFT) method. This perspective is further supported by Han et al.\ \cite{han2024parameter}, who examine PEFT techniques from an LLM perspective. Among these, Low-Rank Adaptation (LoRA) \cite{hu2021loralowrankadaptationlarge} has been widely recognised as an effective method for enhancing LLM performance through adapter-based fine-tuning. These approaches primarily modify newly introduced components while keeping the original large models frozen, thereby reducing computational costs and improving efficiency. However, existing methods operate at the layer level, whereas dropin has the potential to function at the neuron level, akin to neurogenesis in biological brain systems. This distinction presents an open research direction for future exploration.

\begin{table*}[t!]
    \centering
    \renewcommand{\arraystretch}{1.2}
    \begin{tabular}{lcccl}
        \hline
        \textbf{Paper} & \textbf{Year} & \textbf{Neurogenesis} & \textbf{Neuroapoptosis} & \textbf{Comments} \\
        \hline
        Perwej et al.\ \cite{perwej2012neuroplasticity} & 2012 & \checkmark & \checkmark & Start from small or cut from large and early stop \\
        Wagarachchi et al.\ \cite{wagarachchi2017optimization} & 2017 &  & \checkmark & Remove unimportant layers and merge similar neurons \\
        Eriksson et al.\ \cite{eriksson2019dynamic} & 2019 & \checkmark &  & Triggered only when error exceeds threshold \\
        Allam et al.\ \cite{allam2019achieving} & 2019 &  & \checkmark & Application in IoT \\
        Zilly et al. \cite{zilly2021plasticity} & 2021 & \checkmark & \checkmark & Restart frozen weights connection \\
        Camacho et al.\ \cite{camacho2022neuroplasticity} & 2022 &  & \checkmark & Prune entire blocks and add one translator block \\
        Huang et al.\ & 2023 & \checkmark &  & Drop first and grow new neurons \\
        Sowrirajan et al.\ \cite{sowrirajan2024enhancing} & 2024 & \checkmark &  & Triggered only when necessary \\
        Rudroff et al.\ \cite{rudroff2024neuroplasticity} & 2024 &  & \checkmark & Dual learning network inspired by the hippocampus \\
        Steur et al.\ \cite{steur2024step} & 2024 & \checkmark & \checkmark & Skip connection learing \\
        \hline
    \end{tabular}
    \caption{Summary of studies incorporating neurogenesis and neuroapotosis in neural networks.} 
    \label{tab:neuro_table}
\end{table*}

\subsection{Neuroapoptosis: dropout and structural pruning}
Although few AI techniques are directly inspired by neuroapoptosis, those that do exist remain noteworthy. \textbf{NNs can deactivate connections or entirely remove neurons to update their structure}, which we define as neuroapoptosis in artificial neural networks following a technical approximation. It is a process that corresponds to dropout for temporary deactivation and structural pruning for permanent removal.
\subsubsection{Dropout}
Originally, dropout was proposed to prevent overfitting \cite{srivastava2014dropout}, while Baldi et al.\ \cite{baldi2013understanding} provided an alternative perspective, suggesting that dropout can be viewed as a regularisation method applicable to regularised error functions. Wager et al.\ \cite{wager2013dropout} not only regard dropout as a regularisation technique, but also highlight its connection to AdaGrad \cite{duchi2011adaptive}. Gal et al.\ \cite{pmlr-v48-gal16} proposed an alternative perspective, interpreting dropout as a Bayesian approximation for modelling uncertainty, as dropout randomly generates sub-networks during training. However, the most conventional form of dropout, which is commonly referenced, is dropout from neurons: \textbf{a regularisation technique that randomly sets a proportion of units to zero in NNs during training to prevent overfitting and improve generalisation}. The corresponding algorithm is presented in Algorithm \ref{al:dropout}.
\begin{algorithm}
\caption{Dropout Process in NNs}
\label{al:dropout}
\begin{algorithmic}[1]
\REQUIRE Input $\mathbf{x}$, weight matrices $\mathbf{W}^{(l)}$ for each layer $l$, dropout rate $p$
\ENSURE Output $\mathbf{y}$

\STATE \textbf{Training Phase:}
\FOR{each layer $l$}
    \STATE Sample a binary mask $\mathbf{m}^{(l)} \sim \text{Bernoulli}(1 - p)$
    \STATE Apply dropout: $\mathbf{h}^{(l)} = \mathbf{m}^{(l)} \odot f(\mathbf{W}^{(l)} \mathbf{h}^{(l-1)})$ \COMMENT{Element-wise multiplication}
\ENDFOR
\STATE Output $\mathbf{y} = \mathbf{h}^{(L)}$

\STATE \textbf{Inference Phase:}
\FOR{each layer $l$}
    \STATE Compute activations: $\mathbf{h}^{(l)} = f(\mathbf{W}^{(l)} \mathbf{h}^{(l-1)}) \cdot (1 - p)$ \COMMENT{Scale activations}
\ENDFOR
\STATE Output $\mathbf{y} = \mathbf{h}^{(L)}$

\end{algorithmic}
\end{algorithm}
Nevertheless, building on these foundational concepts, several variants of dropout algorithms have been proposed. 

\paragraph{Efficient dropout}To enhance efficiency, Wang et al.\ \cite{wang2013fast} proposed the fast dropout method, which utilises a Gaussian distribution, supported by statistical theory, to compute the expected output. This approach leverages the mean and variance of the dropped-out activations as a deterministic dropout strategy, replacing traditional Monte Carlo sampling. Consequently, it eliminates the need for repeated resampling of neurons and redundant computations, thereby largely reducing computational overhead. Another method, termed Controllable Dropout, was proposed to reduce the memory usage and training time of neural networks \cite{Controll2017}. Unlike traditional dropout, which randomly selects neurons to be dropped, Controllable Dropout deliberately chooses neurons to drop in a way that preserves the regular matrix structure. Specifically, the dropped neurons are grouped in the same rows or columns, thereby enhancing the efficiency of matrix calculations during training. 

\paragraph{Adaptive dropout} The aforementioned dropout strategy can accelerate the training process; however, the dropout rate remains fixed and inflexible. Moreover, determining an appropriate dropout rate requires fine-tuning, making it a challenging hyperparameter to optimise. To address these limitations, Curriculum Dropout \cite{Morerio_2017_ICCV} has been proposed, inspired by learning rate scheduling, to dynamically adjust the dropout rate during training, often leading to improved generalisation. Similarly, Concrete Dropout \cite{gal2017concretedropout} has been introduced to mitigate these challenges by adapting the dropout rate based on the concrete distribution—a continuous relaxation of discrete random variables—while processing incoming data. 

Unlike methods that adaptively adjust the dropout rate during training, Excitation Dropout, inspired by neuroplasticity \cite{zunino2021excitation}, assigns different dropout rates to individual neurons. In this approach, neurons that contribute less to the final output are assigned higher dropout rates. Similarly inspired by brain activity, Continuous Dropout \cite{2018continuousdropout} assigns a continuous probability to each neuron, reflecting the way the brain activates neurons with varying and stochastic probabilities. This probability can be dynamically adjusted during training, allowing for more flexible and biologically motivated regularisation. These adaptive dropout strategies facilitate the automatic determination of an optimal dropout rate, thereby reducing computational costs.

\paragraph{Trainable dropout} Instead of relying on adaptive strategies that involve separate optimisation processes from the neural network training, a more direct approach to dropout with trainable parameters has been proposed. Generalised Dropout \cite{srinivas2016generalizeddropout} introduces a version of dropout, referred to as dropout++, which includes trainable parameters that can be optimised during training. Similarly, Variational Dropout \cite{kingma2015variational} offers a trainable variant of fast dropout, featuring a more flexibly parameterised posterior. These methods provide more advanced, automated dropout techniques, enhancing the generalisability of the models.

\paragraph{Model-specific dropout} Beyond generalised dropout, specialised dropout strategies have been designed to target specific features in certain neural network architectures. Spatial Dropout \cite{Tompson_2015_CVPR} was introduced to drop entire feature maps instead of individual neurons, thereby preserving spatial correlations in convolutional neural networks (CNNs). Similarly, RNN Dropout \cite{semeniuta2016recurrentdropoutmemoryloss} reduces memory consumption by directly dropping recurrent connections in recurrent neural networks (RNNs). A related concept is found in DropConnect \cite{wan2013regularization}, where a subset of network weights, rather than individual neurons, is set to zero. Maxout Dropout \cite{goodfellow2013maxout} was specifically designed for Maxout networks, applying dropout to Maxout units instead of neurons due to the unique structure of this architecture. 

Given the diversity of neural network architectures, a wide range of tailored dropout strategies have been developed to optimise performance, ensuring that dropout mechanisms align with the structural and functional characteristics of different models.

\paragraph{Recent dropout} Recently, dropout strategies have continued to evolve alongside the rapid advancements in NNs architectures. Some approaches have adapted existing dropout techniques to state-of-the-art models. For instance, dynamic dropout has been applied to Transformers \cite{altarabichi2024dropkan}, improving training efficiency, while diffusion dropout \cite{tai2024dose} has been introduced to encourage models to prioritise conditioning factors in speech generation. Additionally, LoRA Dropout \cite{lin2024lora} has been proposed to mitigate overfitting in LLMs.

Other research efforts have focused on extending and refining previous dropout strategies. R-Drop \cite{liang2021rdropregularizeddropoutneural} modifies traditional dropout by enforcing consistency between different dropout-applied versions of a neural network, thereby enhancing regularisation. Inspired by plasticity loss, Activation by Interval-wise Dropout \cite{park2025activation} introduces an adaptive dropout mechanism that assigns different dropout rates to different preactivation intervals.

These recent developments indicate a growing interest in dropout research, summarised in Table \ref{tab:dropout_methods}, with ongoing exploration of biologically inspired mechanisms that could further enhance the effectiveness of dropout strategies.
\begin{table*}[t!]
    \renewcommand{\arraystretch}{1.2}

    \centering
    \begin{tabular}{lll}
        \toprule
        \textbf{Paper} & \textbf{Year} & \textbf{Comments} \\
        \midrule
        Fast Dropout \cite{wang2013fast} & 2013 & Deterministic dropout strategy to replace traditional Monte Carlo sampling \\
        DropConnect \cite{wan2013regularization} & 2013 & Dropout weights \\
        Maxout Dropout \cite{goodfellow2013maxout} & 2013 & Dropout maxout unit \\
        Variational Dropout \cite{kingma2015variational} & 2015 & Parameterised posterior of fast dropout \\
        Spatial Dropout \cite{Tompson_2015_CVPR} & 2015 & CNN dropout \\
        Generalised Dropout \cite{srinivas2016generalizeddropout} & 2016 & Dropout with learnt weights \\
        RNN Dropout \cite{semeniuta2016recurrentdropoutmemoryloss} & 2016 & RNN dropout to save memory \\
        Controllable Dropout \cite{Controll2017} & 2017 & Dropout to preserve matrix structure \\
        Curriculum Dropout \cite{Morerio_2017_ICCV} & 2017 & Scheduling dropout rate \\
        Concrete Dropout \cite{gal2017concretedropout} & 2017 & Adaptive rate based on concrete distribution \\
        Continuous Dropout \cite{2018continuousdropout} & 2018 & Continuous probability of dropout \\
        Information Dropout \cite{achille2019information} & 2018 & Dropout based on Information such as feature map \\
        Weight Dropout \cite{sanjar2020weight} & 2020 & Dropout weights directly \\
        Excitation Dropout \cite{zunino2021excitation} & 2021 & Dropout based on influence for final results \\
        R-Drop \cite{liang2021rdropregularizeddropoutneural} & 2021 & Regularised dropout \\
        Dropout KAN \cite{altarabichi2024dropkan} & 2024 & Dropout post-activation for KANs \\
        Transformer Dropout \cite{altarabichi2024dropkan} & 2024 & Improve training efficiency for Transformer \\
        Diffusion Dropout \cite{tai2024dose} & 2024 & Add training conditions for diffusion \\
        LoRA Dropout \cite{lin2024lora} & 2024 & LLMs mitigation \\
        Activation by Interval-wise Dropout \cite{park2025activation} & 2025 & Inspired by plasticity loss, dropout in preactivation intervals \\
        \bottomrule
    \end{tabular}
    \caption{Overview of different dropout techniques}
    \label{tab:dropout_methods}
\end{table*}

\paragraph{Weights dropout}
Other than dropout on neurons, previous research has put some efforts on weights dropout. Dropout KAN \cite{altarabichi2024dropkan} was developed for Kolmogorov–Arnold Networks (KANs), where learning is based on kernel functions rather than traditional weight-based updates, therefore its dropout operates on post-activations. Weight Dropout, introduced by Sanjar et al. \cite{sanjar2020weight}, incorporates a dropout mask applied directly to the weights. Their results demonstrate that this method yields improved performance across various tasks.

\paragraph{Information dropout}
Beyond focusing solely on dropout in NNs, prior studies have explored broader aspects of learning dynamics in terms of information. One of the earliest discussions appears in \cite{achille2018critical}, which introduces the concept of a critical learning period in deep NNs. This work suggests that, during the initial phase of training, networks experience a surge in information gain, which subsequently diminishes as training progresses. Furthermore, it emphasises the importance of information forgetting as an integral part of the learning process. Building upon this notion, Kleinman et al.\ \cite{kleinman2023critical} argue that multisensory information is progressively integrated within NNs throughout training—a process that exemplifies the model’s capacity for information acquisition. They also maintain that the critical learning period persists in this context.

Therefore, controlling the amount of information at different stages of training plays a crucial role in the training of neural networks. Achille et al.\ \cite{achille2019information} propose a framework for quantifying the complexity of task-relevant information. Based on information-theoretic principles, they introduce Information Dropout, a technique that selectively drops information—such as elements in feature maps—during training. This approach allows the model to automatically adapt to the data and more effectively utilize architectures with limited capacity \cite{achille2018information}.

\subsubsection{Structual pruning}

Unlike dropout, which does not permanently terminate neurons, structural pruning removes neurons or connections that are no longer useful, allowing them to be eliminated entirely. Structural pruning is defined as the process of reducing a network’s size by removing parameters \cite{blalock2020state}, thereby improving efficiency while maintaining performance.

\paragraph{Pruning types}
Pruning can be applied at different levels within a neural network. At the lowest level, weight pruning \cite{han2015learningweightsconnectionsefficient} employs local and iterative pruning to remove individual connections. Moving to a higher level, neuron pruning techniques, such as those using the Polarization Regularizer \cite{zhuang2020neuron} or Importance Score Propagation \cite{Yu_2018_CVPR}, selectively eliminate entire neurons, though the most traditional approach remains pruning neurons whose weights are zero \cite{bondarenko2015neurons}. At an even higher level, layer pruning \cite{guenter2024concurrenttraininglayerpruning} reduces the number of layers, easing the parallelization of sequential computations. Additionally, the Centered Kernel Alignment (CKA) metric has been proposed to assess the relevance of layers for pruning \cite{pons2024effectivelayerpruningsimilarity}. While pruning significantly improves computational efficiency by removing redundant or unnecessary connections proved by more literature references, improper selection of pruning criteria may lead to a decline in model performance. Additionally, Wagarachchi et al.\  \cite{wagarachchi2017optimization} propose an alternative pruning strategy by removing layers deemed unimportant (i.\,e., those with minimal influence on the output) and merging similar neurons, resulting in performance improvements across six datasets. 

\paragraph{Pruning criteria and models}
Determining the optimal pruning criterion is challenging, as no universal best criterion exists due to the diversity of model architectures, applications, and training strategies. However, several effective approaches have been proposed.

One widely used method is magnitude pruning, which removes weights smaller than a predefined threshold, as they have minimal impact on neural network performance. An example is Layer-Adaptive Magnitude-based Pruning \cite{lee2020layer}. A survey \cite{gupta2024complexity} further discusses its effectiveness compared to more complex pruning algorithms. Another approach is activation-based pruning, which eliminates neurons that consistently exhibit low activation levels. Experiments reported in \cite{ardakani2017activation} demonstrated performance improvements using this technique. Ganguli et al.\ \cite{ganguli2024activation} further explored this strategy, suggesting that pruning can function as both a regularization and dimensionality reduction technique. This approach has also been applied in various contexts, such as model safety \cite{dhillon2018stochastic}, CNN acceleration \cite{ye2020accelerating}, and attention-based model pruning \cite{samal2020attention}.

Additionally, gradient-based pruning identifies redundant neurons by removing those with the smallest gradients concerning the final loss, as these neurons contribute minimally to model learning. A gradient-based framework for analysing neural network pruning has been proposed \cite{lubana2020gradient}, and this strategy has also been shown to be effective for fine-tuning \cite{cai2022prior}. Applications of gradient pruning extend to acceleration \cite{mcdanel2022accelerating} and improving fairness \cite{lin2022fairgrape}.

Other formats of pruning also exist. For example, Camacho et al.\ \cite{camacho2022neuroplasticity} simulate induced injuries in the network by pruning entire convolutional layers or blocks. To maintain structural integrity, a translator block is introduced to adjust the dimensionality before retraining the network. This method has been successfully applied to high-performing architectures such as VGG16 \cite{simonyan2015deepconvolutionalnetworkslargescale} and MobileNets \cite{howard2017mobilenetsefficientconvolutionalneural}, largely reducing parameter counts while preserving performance. 

Overall, different methods aim to identify the least contributive neurons through different criteria, enabling improvements in efficiency, reduced complexity, and, in some cases, enhanced model performance. 

Structural pruning exhibits similar effectiveness to dropout, but with a key distinction: pruned neurons and connections are permanently removed and cannot be reinstated. Consequently, structural pruning offers more pronounced benefits in terms of memory efficiency and computational savings. However, excessive pruning can lead to Layer Collapse \cite{shabgahi2023layercollapse}, where over-pruning results in the inability to fully learn features, leading to information loss. Interestingly, some studies leverage this phenomenon as an indicator that structural pruning has reached its optimal and maximum extent.

Once pruning is completed, the neural network must undergo retraining to reallocate and adjust the remaining neurons' weights. This can be achieved through continuous learning or knowledge distillation, as discussed earlier, ensuring that the model retains its learning capacity despite structural modifications.

\subsection{Neuroplasticity: dynamical adjustment of neurons and their connection}
\begin{figure*}
    \centering
    \includegraphics[width=0.8\linewidth]{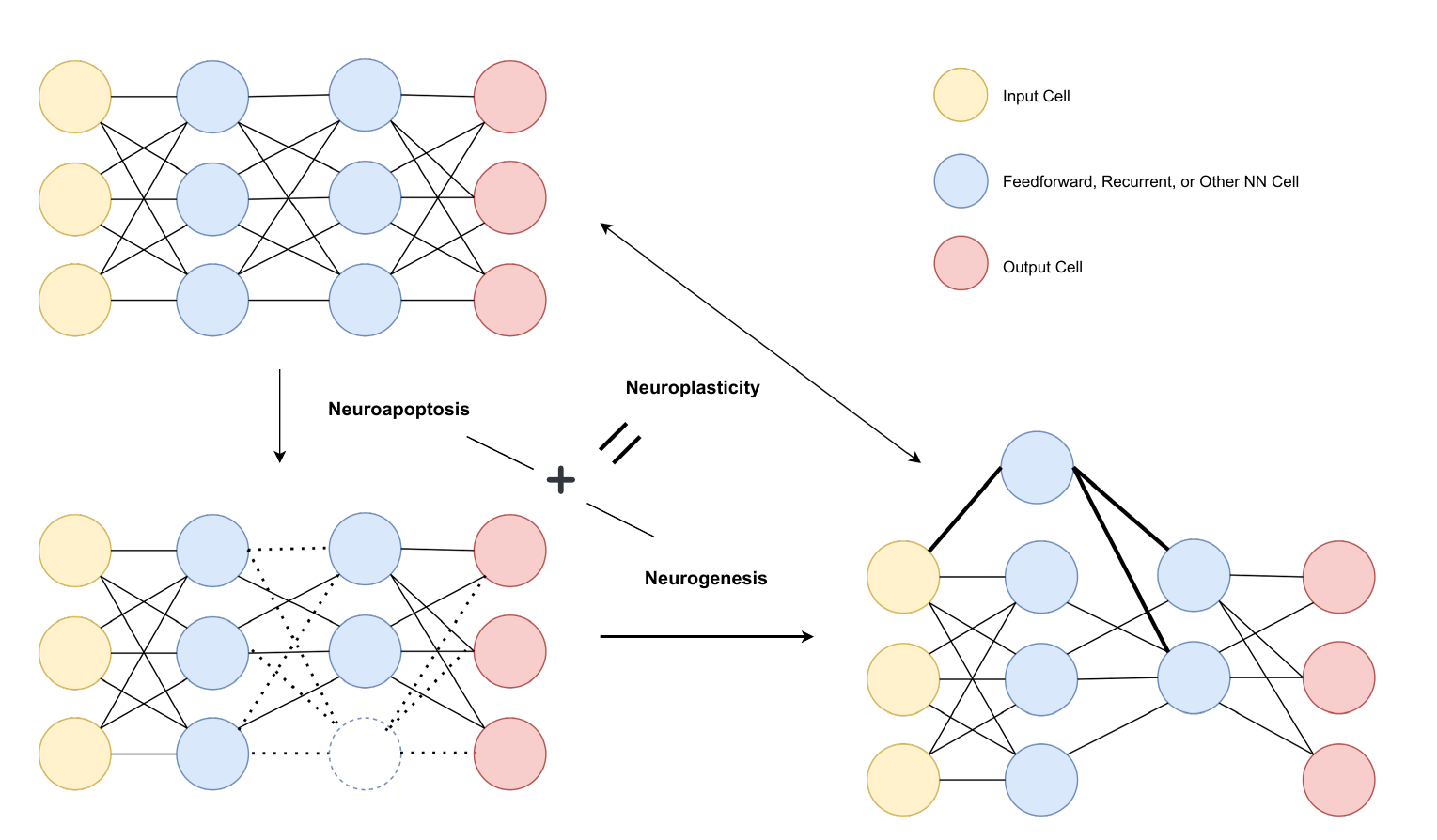}
    \caption{Neuroplasticity, neurogenesis, and neuroapoptosis relationship. Neuroplasticity is the comprehensive mechanism encompassing both neurogenesis and neuroapoptosis.}
    \label{fig:logistics}
\end{figure*}

This mechanism of neuroplasticity closely resembles biological neurons, which adjust their structure in response to external or internal stimuli. Although in huamn brain neuroplasticity, neurogenesis, and neuroapoptosis follow distinct processes, the combined effects of neurogenesis and neuroapoptosis contribute to both the addition and removal of neurons, a phenomenon reflected in neuroplasticity as discussed previously. \textbf{Therefore, we define artificial NNs neuroplasticity as the sum of artificial neural network neurogenesis and neuroapoptosis, facilitated by technical approximation.}
The relationship between the three brain activities is shown in Figure \ref{fig:logistics}.
\begin{algorithm}[ht!]
\caption{Neuroplasticity in NNs}
\label{al:neuroplasticity}
\begin{algorithmic}[1]
\REQUIRE Input $\mathbf{x}$, weight matrices $\mathbf{W}^{(l)}$ for each layer $l$
\ENSURE Output $\mathbf{y}$

\STATE \textbf{Training Phase:}
\FOR{New data \OR $\Delta \max(\text{loss}_{epoch}, \text{loss}_{epoch-1}) < \delta$}

\item  \COMMENT{The layer with the largest LRP value exceeds the threshold defined by the average LRP value.}
        \STATE Sample a number of new neurons and connections for layer $i$ $\mathbf{W}^{(l)_{new}}$ \COMMENT{Create additional neurons}
        \STATE Compute activations:

        $\mathbf{h}^{(l-1)_{new}} = f(\text{concat}(\mathbf{W}^{(l-1)},\mathbf{W}^{(l-1)_{new}})  \mathbf{h}^{(l-2)})$

        $\mathbf{h}^{(l)} = f(\text{concat}(\mathbf{W}^{(l)},\mathbf{W}^{(l)_{new}})  \mathbf{h}^{(l-1)_{new}})$ \COMMENT{Two consecutive layers will be influenced as the shape of weight matrices has been changed}

    \IF{$Criterion(val)$ decrease}
    \STATE \textbf{break} 
\ENDIF
\ENDFOR

    \IF{$Criterion(val)$ decrease}
        \STATE Pick those useless weights position and create binary mask $\mathbf{m}^{(l)}$ \COMMENT{Useless weights position could be acquired by the same LPR algorithm}
        \STATE Apply dropout: $\mathbf{h}^{(l)} = \mathbf{m}^{(l)} \odot f(\mathbf{W}^{(l)} \mathbf{h}^{(l-1)})$ \COMMENT{Element-wise multiplication}
    \ELSE
        \STATE Compute activations: $\mathbf{h}^{(l)} = f(\mathbf{W}^{(l)} \mathbf{h}^{(l-1)})$ 
    \ENDIF

\STATE Output $\mathbf{y} = \mathbf{h}^{(L)}$

\STATE \textbf{Inference Phase:}
\FOR{each layer $l$}
    \STATE Compute activations: $\mathbf{h}^{(l)} = f(\mathbf{W}^{(l)} \mathbf{h}^{(l-1)})$ \COMMENT{Use all neurons during inference}
\ENDFOR
\STATE Output $\mathbf{y} = \mathbf{h}^{(L)}$

\end{algorithmic}
\end{algorithm}

The suggested approach further shows some similarities with concepts introduced in the literature: Perwej et al.\ \cite{perwej2012neuroplasticity} propose a neuroplasticity-inspired approach in NNs to accelerate training and enhance model performance. Their method involves initially utilising a small network and progressively adding neurons until the AI successfully learns a specific skill, demonstrating neuron addition. Subsequently, they implement a neuron removal process by starting with a large network and gradually eliminating connections while ensuring the model retains its essential skills. Experimental results indicate that pruning two thirds of the connections in a Multilayer Perceptron (MLP) does not considerably degrade performance. Steur et al.\ \cite{steur2024step} propose an additional method that enables Residual Networks and Highway Networks to establish skip connections using Adaptive Nonlinearity Gates (ANGs). Rudroff et al.\ \cite{rudroff2024neuroplasticity} clearly discuss how the biological system’s ability to balance rapid learning with long-term memory retention can inspire novel AI architectures. They propose a dual-learning architecture: a fast-learning module in the buffer or short-term memory, coupled with offline consolidation replay to the knowledge base or long-term memory. This approach draws inspiration from brain activity during Non-REM (Non-Rapid Eye Movement) and REM sleep, incorporating mechanisms such as SWRs (Sharp Wave Ripples) for reactivation and strengthening, and BARRs (Bidirectional Associative Recall Responses). They present testable hypotheses for both human cognition and neural networks, further bridging AI and neuroscience. Moreover, Zilly et al. \cite{zilly2021plasticity} observed that weights in neural networks tend to become mutually frozen during sequential task learning, highlighting the limitations of this learning paradigm. To address this issue, they propose reinitializing the connections in the frozen parts of the network, thereby enhancing the plasticity of neural networks and improving their ability to transfer knowledge across tasks. These concepts originated from neuroplasticity, which focuses on pruning neurons deemed redundant or non-contributory to the model's performance. In contrast, approaches inspired by neurogenesis emphasise the activation of new neurons to enhance the model’s overall performances.

Also, the aforementioned multiple dropin-like strategies could also be combined with dropout to enhance the flexibility and comprehensiveness of the overall approach for dynamic neuronal modifications during training. This mechanism would be analogous to neuroplasticity in brain activity. We propose a corresponding algorithms in Algorithm \ref{al:neuroplasticity}.

In AI, the addition and removal of neurons in NNs can be considered analogous to neuroplasticity. Recent studies indicate that drawing inspiration from neurogenesis and neuroplasticity can enhance both the performance and efficiency of neural networks. These contributions are summarised in Table \ref{tab:neuro_table}.

\section{Implications and challenges}
\label{sec3}
There are numerous implications of this research based on prior studies. 

First, neuroscience has the potential to advance AI by opening new research avenues \cite{hassabis2017neuroscience}. As discussed, existing NN dropout strategies are widely adopted, with some algorithms drawing inspiration from neurogenesis and neuroplasticity. As AI enables a deeper understanding of neuronal activity, and as our knowledge in this area expands, new directions may emerge for developing novel NN architectures. This could include innovative dropout and dropin strategies informed by neuronal activity. 

Existing technologies have been developed to simulate neural networks in the brain \cite{rinke2018scalable}. For instance, algorithms have been introduced to mimic neuronal plasticity, providing deeper biological insights into brain function. These advancements not only enhance our understanding of the brain but also foster potential AI innovations in a continuous cycle of inspiration.

The advantages of these approaches inspired by biology are evident. The incorporation of neurogenesis can enhance a model’s capacity for learning, while neuroapoptosis strategies contribute to reducing computational costs and mitigating the risks of overfitting or underfitting. By selectively adding or ceasing the training of certain neurons in later stages, these techniques facilitate more efficient resource allocation and adaptive model optimisation. Moreover, this approach holds great potential for improving performance in the era of LLMs, as retaining more meaningful and non-redundant neurons enhances a model's capacity to acquire and represent knowledge effectively.

However, this direction is not without challenges. Since dropin and dropout operate at the neuronal level, identifying the root cause of errors may be difficult, as such modifications occur at a micro scale. 
Nevertheless, we advocate for further exploration and experimentation in this area, as prior research suggests its potential to yield tremendous advancements.

The second implication is the more profound potential for human-machine integration. The ability to understand neuronal activity suggests that AI could be developed based on human neural processes, leading to the emergence of new applications as mutual understanding between humans and AI deepens. As this understanding expands, additional topics will emerge, for instance in areas where brain data is combined with AI-powered storage solutions. This ambitious field remains ripe for further exploration and discovery.

\section{Conclusion}
In this paper, we have reviewed how AI techniques can be inspired by biological processes, such as neurogenesis and neuroplasticity. We also examined corresponding strategies, such as dropout, which has demonstrated important performance in recent years. In contrast to dropout, we introduced the concept of dropin and the combination of dropin and dropout in a dynamic ongoing neuroplastic behaviour cycle and discussed its potential in the era of AI, its subfield deep learning, and in particular for the present and oncoming large models that continually learn and evolve. Furthermore, we explored the mutual development between AI and biology, analysing the implications of this evolution. We advocate for increased research efforts in this area, emphasising the need for practical experimentation and empirical validation of the proposed algorithms to substantiate their effectiveness.

\sloppy
\bibliographystyle{IEEEtran}
\bibliography{reference}

\end{document}